\documentclass[runningheads]{llncs}
\usepackage{graphicx}
\usepackage{comment}
\usepackage{amsmath,amssymb} 
\usepackage{color}
\usepackage{subfigure}
\usepackage{mathrsfs} 
\usepackage{bm}
\usepackage{mdwlist}
\usepackage{array,multirow}
\usepackage{paralist}
\usepackage{bbding}
\usepackage{enumerate}
\usepackage{booktabs}
\usepackage{url}
\usepackage{caption}
\usepackage{xspace}
\usepackage{supertabular}
\usepackage{threeparttable}

\def\eg{{\em e.g.}}
\def\ie{{\em i.e.}}
\def\etal{{\em et al.}}

\begin{document}
\pagestyle{headings}
\mainmatter
\def\ECCVSubNumber{1951}  

\title{Spatial Attention Pyramid Network for Unsupervised Domain Adaptation} 

\titlerunning{Spatial Attention Pyramid Network for Unsupervised Domain Adaptation}
%
\authorrunning{C. Li et al.}

\newcommand*\samethanks[1][\value{footnote}]{\footnotemark[#1]}
\author{Congcong Li\inst{1}\and
Dawei Du\inst{2}\and
Libo Zhang\inst{3}$^,$\thanks{Corresponding author (libo@iscas.ac.cn).}\and
Longyin Wen\inst{4}\and
Tiejian Luo\inst{1}$^,$\samethanks\and
Yanjun Wu\inst{3}\and
Pengfei Zhu\inst{5}}
\institute{University of Chinese Academy of Sciences, Beijing, China \and
University at Albany, State University of New York, Albany, NY, USA \and
State Key Laboratory of Computer Science, ISCAS, Beijing, China \and
JD Finance America Corporation, Mountain View, CA, USA \and
Tianjin University, Tianjin, China}

\maketitle

\begin{abstract}
Unsupervised domain adaptation is critical in various computer vision tasks, such as object detection, instance segmentation, and semantic segmentation, which aims to alleviate performance degradation caused by domain-shift. Most of previous methods rely on a single-mode distribution of source and target domains to align them with adversarial learning, leading to inferior results in various scenarios. To that end, in this paper, we design a new spatial attention pyramid network for unsupervised domain adaptation. Specifically, we first build the spatial pyramid representation to capture context information of objects at different scales. Guided by the task-specific information, we combine the dense global structure representation and local texture patterns at each spatial location effectively using the spatial attention mechanism. In this way, the network is enforced to focus on the discriminative regions with context information for domain adaptation. We conduct extensive experiments on various challenging datasets for unsupervised domain adaptation on object detection, instance segmentation, and semantic segmentation, which demonstrates that our method performs favorably against the state-of-the-art methods by a large margin. Our source code is available at \url{https://isrc.iscas.ac.cn/gitlab/research/domain-adaption}.
\keywords{Unsupervised Domain Adaptation $\cdot$ Spatial Attention Pyramid $\cdot$ Object Detection $\cdot$ Semantic Segmentation $\cdot$ Instance Segmentation}
\end{abstract}

\section{Introduction}
Over the past few years, deep neural network (DNN) significantly pushed forward the state-of-the-art in several tasks in computer vision field, such as object detection \cite{DBLP:journals/pami/RenHG017,DBLP:conf/cvpr/ZhangWBLL18}, instance segmentation \cite{DBLP:conf/iccv/HeGDG17,DBLP:conf/cvpr/HayderHS17}, and semantic segmentation \cite{DBLP:conf/cvpr/LongSD15,DBLP:journals/corr/ChenPSA17}. Notably, the DNN-based methods rely on large-scale annotated training data, which is difficult to cover diverse application domains. That is, the feature distributions (\eg, local texture, object appearance and global structure) between source domain and target domain are dissimilar or even completely different. To avoid expensive and time-consuming human annotation, unsupervised domain adaptation is proposed to learn discriminative cross-domain representation in such domain shift circumstance \cite{quinonero2008covariate}.

Most of previous methods \cite{DBLP:journals/corr/HoffmanWYD16,DBLP:conf/nips/LiuBK17,DBLP:conf/icml/HoffmanTPZISED18} attempt to globally align the entire distributions between the source and target domains. However, it is challenging to generate a unified adaptation function for various scene layouts and appearance variation of different objects. Recent methods focus on transferring texture and color statistics within object instances or local patches. To deal with domain adaptation in the object detection and instance segmentation tasks, the basic idea in \cite{DBLP:conf/cvpr/Chen0SDG18,DBLP:conf/cvpr/ZhuPYSL19} is to exploit discriminative features in bounding boxes of objects and attempt to align them across both source and target domains. However, the context information around the objects is not fully exploited, causing inferior results in some scenarios. Meanwhile, some domain adaptation methods for semantic segmentation \cite{DBLP:journals/corr/abs-1901-05427,DBLP:conf/cvpr/Luo0GYY19} enforce the semantic consistency between the pixels or local patches of the two domains, leading to deficiencies of critical information from object-level patterns. To that end, recent methods \cite{DBLP:conf/cvpr/SaitoUHS19,DBLP:journals/corr/abs-1911-02559} concatenate global context feature and instance-level feature for distribution alignment, and optimize the model based on several loss terms for global-level and local-level features with pre-set weights. However, this method fails to exploit the context information of objects, which is not optimal in challenging scenarios.

In this paper, we design the spatial attention pyramid network (SAPNet) to solve unsupervised domain adaptation for object detection, instance segmentation, and semantic segmentation. Inspired by spatial pyramid pooling \cite{DBLP:journals/pami/HeZR015}, we construct the spatial pyramid representation with multi-scale feature maps, which integrates full of holistic (image-level) and local (regions of interest) semantic information. Meanwhile, we design a task-specific guided spatial attention mechanism to capture multi-scale context information. In this way, discriminative semantic regions are attended in a soft manner to extract features for adversarial learning. Extensive experiments are conducted on various challenging domain-shift datasets, such as Cityscapes \cite{DBLP:conf/cvpr/CordtsORREBFRS16} to FoggyCityscapes \cite{DBLP:journals/ijcv/SakaridisDG18}, PASCAL VOC \cite{DBLP:journals/ijcv/EveringhamGWWZ10} to Clipart \cite{DBLP:conf/cvpr/InoueFYA18}, and GTA5 \cite{DBLP:conf/eccv/RichterVRK16} to Cityscapes \cite{DBLP:conf/cvpr/CordtsORREBFRS16}. It is worth mentioning that the proposed method surpasses the state-of-the-art methods on various tasks, \ie, object detection, instance segmentation, and semantic segmentation. For example, our SAPNet improves from Cityscapes \cite{DBLP:conf/cvpr/CordtsORREBFRS16} to FoggyCityscapes \cite{DBLP:journals/ijcv/SakaridisDG18} by $3\%$ mAP in terms of object detection, and achieves comparable accuracy from GTA5 \cite{DBLP:conf/eccv/RichterVRK16} to Cityscapes \cite{DBLP:conf/cvpr/CordtsORREBFRS16} in terms of semantic segmentation.

{\flushleft \textbf{Contributions}}. 1) We propose a new spatial attention pyramid network to solve the unsupervised domain adaptation task for object detection, instance segmentation, and semantic segmentation. 2) We develop a task-specific guided spatial attention pyramid learning strategy to merge multi-level semantic information in feature maps of the pyramid. 3) Extensive experiments conducted on various challenging domain-shift datasets for object detection, instance segmentation, and semantic segmentation, demonstrate the effectiveness of the proposed method, surpassing the state-of-the-art methods.

\section{Related Works}
{\noindent \textbf{Unsupervised domain adaptation.}} Several methods have been proposed for unsupervised domain adaptation in terms of several tasks such as object detection \cite{DBLP:conf/cvpr/Chen0SDG18,DBLP:conf/cvpr/SaitoUHS19,DBLP:journals/corr/abs-1911-02559}, instance segmentation \cite{DBLP:conf/cvpr/ZhuPYSL19} and semantic segmentation~\cite{DBLP:journals/corr/HoffmanWYD16,DBLP:journals/corr/abs-1901-05427,DBLP:conf/cvpr/Luo0GYY19}. For object detection domain adaptation, Chen~\etal~ \cite{DBLP:conf/cvpr/Chen0SDG18} align source and target domain both on image level and instance level using the gradient reverse layer \cite{DBLP:conf/icml/GaninL15}. Zhu~\etal~\cite{DBLP:conf/cvpr/ZhuPYSL19} mine the discriminative regions (pertinent to object detection) using $k$-means clustering and align them across both domains, which is applied in object detection and instance segmentation. Recently, the strong-weak adaptation method is proposed in~\cite{DBLP:conf/cvpr/SaitoUHS19}. It focuses the adversarial alignment loss toward images that are globally similar, and away from images that are globally dissimilar by employing focal loss \cite{DBLP:conf/iccv/LinGGHD17}. Shen~\etal~\cite{DBLP:journals/corr/abs-1911-02559} propose a gradient detach based stacked complementary losses method that adapt source domain and target domain on multiple layers. On the other hand, Hoffman~\etal~\cite{DBLP:journals/corr/HoffmanWYD16} perform global domain alignment in a novel semantic segmentation network with fully convolutional domain adversarial learning. Tsai~\etal~\cite{DBLP:journals/corr/abs-1901-05427} learn discriminative feature representations of patches in the source domain by discovering multiple modes of patch-wise output distribution through the construction of a clustered space. Luo~\etal~\cite{DBLP:conf/cvpr/Luo0GYY19} introduce a category-level adversarial network to enforce local semantic consistency on the output space using two distinct classifiers. However, the aforementioned methods only consider domain adaptation in two levels, \ie, aligning the feature maps of the whole image or local regions with a fixed scale. Different from them, we design the spatial pyramid representation to capture multi-level semantic patterns within the image for better adaptation between the source domain and target domain.

{\noindent \textbf{Attention mechanism.}} To focus on the most discriminative features, various attention mechanisms have been explored. SENet \cite{DBLP:conf/cvpr/HuSS18} develops the Squeeze-and-Excitation (SE) block that adaptively recalibrates channel-wise feature responses. Non-local networks \cite{DBLP:conf/cvpr/0004GGH18} capture long-range dependencies by computing the response at a position as a weighted sum of the features at all positions in the input feature maps. SKNet \cite{DBLP:conf/cvpr/LiW0019} uses softmax attention to fuse multiple feature maps of different kernel sizes in a weighted manner to adaptively adjust the receptive field size of the input feature map. Except channel-wise attention, CBAM \cite{DBLP:conf/eccv/WooPLK18} introduce spatial attention by calculating the inter-spatial relation of features. To highlight transferable regions in domain adaptation, Wang~\etal~\cite{DBLP:conf/aaai/WangLYLW19} use multiple region-level domain discriminators and single image-level domain discriminator to generate transferable local and global attention, respectively. Sun and Wu~\cite{DBLP:journals/corr/abs-1901-06322} integrates atrous spatial pyramid, cascade attention mechanism and residual connections for image synthesis and image-to-image translation. As previous works \cite{DBLP:conf/cvpr/SinghD18,DBLP:conf/cvpr/LinDGHHB17} have shown the importance of multi-scale information, we propose the attention pyramid learning to better adapt the source domain and target domain. Specifically, we employ the task-specific information to guide pyramid attention to make full use of semantic information of different feature maps at different levels.

{\noindent \textbf{Detection and segmentation networks.}} The performance of object detection and segmentation has boosted with the development of deep convolutional neural networks. Faster R-CNN \cite{DBLP:journals/pami/RenHG017} is an object detection framework that predicts class-agnostic coarse object proposals using a region proposal network (RPN) and then extract fix-sized object features to classify object category and refine object location. Moreover, He~\etal~\cite{DBLP:conf/iccv/HeGDG17} extend Faster R-CNN by adding a branch for predicting instance segmentation results. For semantic segmentation, the DeepLab-v2 method~\cite{DBLP:journals/pami/ChenPKMY18} develops atrous spatial pyramid pooling (ASPP) modules to segment objects at multiple scales. For fair comparison, we propose the spatial attention pyramid network based on the same detection and segmentation frameworks as that in the previous domain adaptation methods.

\section{Spatial Attention Pyramid Network}
\label{sec:method}
\begin{figure}[t]
\centering
\includegraphics[width=\textwidth]{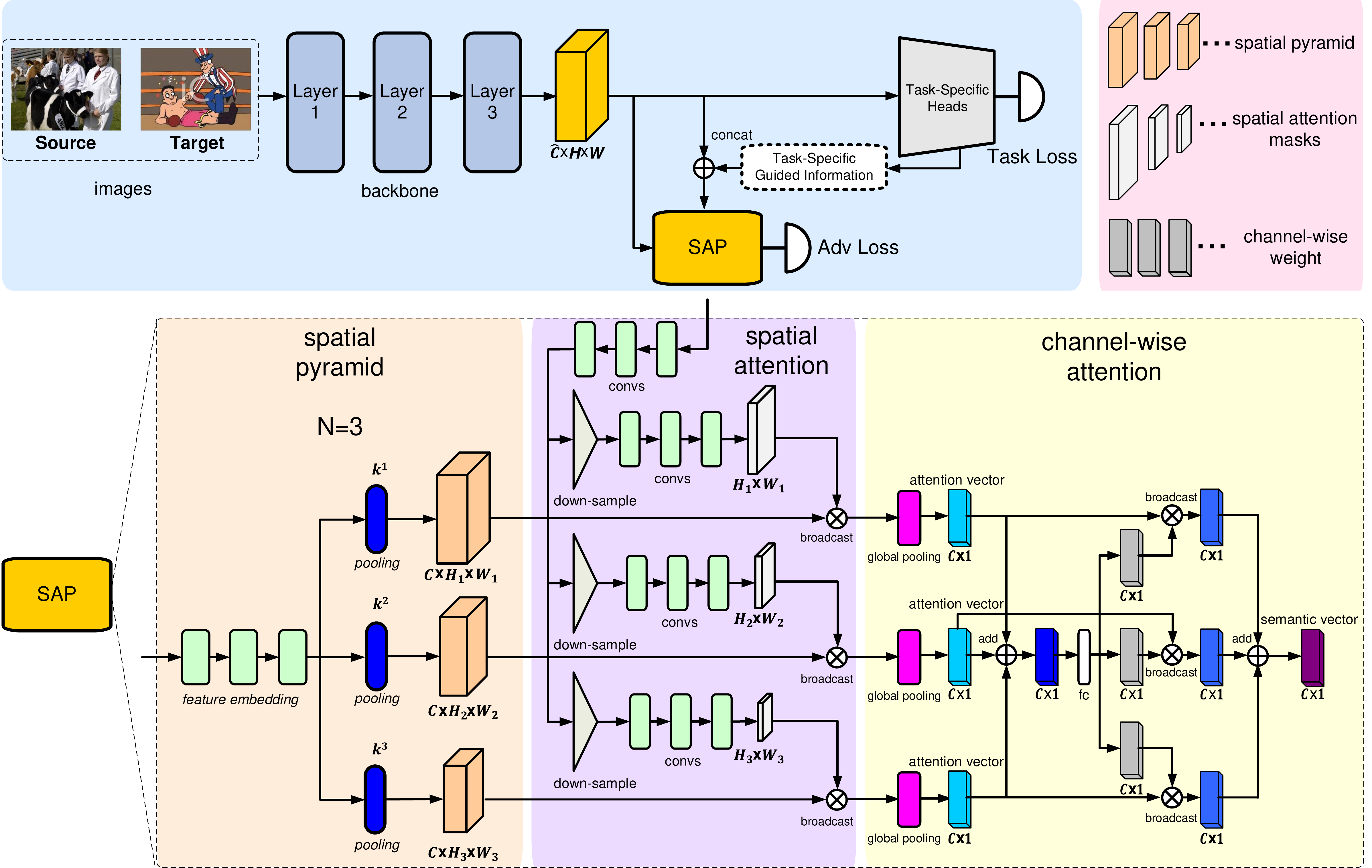}
\caption{The framework of spatial attention pyramid network. For clarity, we only show $N=3$ levels in the spatial pyramid.}
\label{fig:framework}
\end{figure}

We design a spatial attention pyramid network (SAPNet) to solve various computer vision tasks, such as object detection, instance segmentation, and semantic segmentation. First of all, we define the labeled source domain $\mathcal{D}_s$ and the unlabeled target domain $\mathcal{D}_t$, which are subject to the complex and unknown distributions in the source and target domains. Our goal is to find discriminative representation for the distributions of both source and target domains to capture various semantic information and local patterns in different tasks. The architecture of SAPNet is presented in Fig. \ref{fig:framework}.

{\flushleft {\bf Spatial pyramid representation}.}
According to \cite{DBLP:journals/pami/HeZR015,DBLP:conf/cvpr/LazebnikSP06}, spatial pyramid pooling can maintain spatial information by pooling in local spatial bins. To better adapt source and target domains, we develop a spatial pyramid representation to exploit the underlying distribution within an image.

Specifically, as shown in Fig. \ref{fig:framework}, the feature map $\hat{f} \in \mathbb{R}^{\hat{C} \times H \times W}$ is extracted from the backbone $G$ of the network, where $\hat{C}$, $H$ and $W$ are the channel dimension, height, width of feature maps respectively. To improve efficiency, we first reduce the number of channels in $\hat{f}$ to $\bar{f} \in \mathbb{R}^{C \times H \times W}$ gradually by using three $1\times1$ convolutional layers, \ie, we set $C = 256$ in all our experiments. Second, we use multiple average pooling layers with different sizes to operate upon the feature map $\bar{f}$ separately. The sizes of the pooling operation are $\{k^{n}\}_{n=1}^{N}$, where $N$ is the number of pooling layers. That is, the rectangular pooling region with the size $k^n$ at each location of $\bar{f}$ is down-sampled to the average value of each region, resulting in the pyramid of $N$ pooled feature maps $\{f^1, \dots, f^N\}$. In this way, every pooled feature map $f^n \in \mathbb{R}^{C \times H_{n} \times W_{n}}$ in the pyramid can encode different semantic information of objects or layouts within the image.

It is worth mentioning that the proposed spatial pyramid representation is related with spatial pyramid pooling (SPP) for visual recognition \cite{DBLP:journals/pami/HeZR015}. While they share the pooling concept, we would like to highlight two important differences. First, we use average pooling instead of max pooling to construct the spatial pyramid representation. It can better capture the overall strength of local patterns such as edges and textures, which is demonstrated in the ablation study. Second, SPP pools the features with just a few window sizes and concatenates them to generate fixed-length representation; while SAP is designed to capture multi-scale context information of all levels in the pyramid. Thus, it is difficult to use a large number of windows with different sizes for SPP due to computational complexity.

{\flushleft {\bf Attention mechanism.}}
Moreover, we integrate the spatial attention pyramid strategy to enforce the network to focus on the most discriminative semantic regions and feature maps. There are mainly two advantages of introducing the attention mechanism in the spatial pyramid architecture. First, there exist different local patterns in each spatial location of feature maps. Second, different feature maps in the pyramid have different contributions to the semantic representation. The detailed learning method is described in two aspects as follows.

To facilitate highlighting the most discriminative semantic regions, the spatial attention masks for the pyramid $\{f^1,\dots,f^N\}$ are learned based on the guided information from the task-specific heads (\ie, object detection, instance segmentation and semantic segmentation). For object detection and instance segmentation, the guided information is the output map with the size of $A\times H \times W$ from the classification head of region proposal network (RPN). It can predict object's confidence in terms of the locations in feature maps for all the number of anchors $A$. That is, it encodes the distribution of objects which is suitable for object detection and instance segmentation problem. For semantic segmentation, the guided information is the output map with the size of $C_{sem}\times H \times W$ from the segmentation head, where $C_{sem}$ is the number of semantic categories. We denote the guided map as $\mathcal{\hat{P}}$.

\begin{figure}[t]
\centering
\includegraphics[width=.75\linewidth]{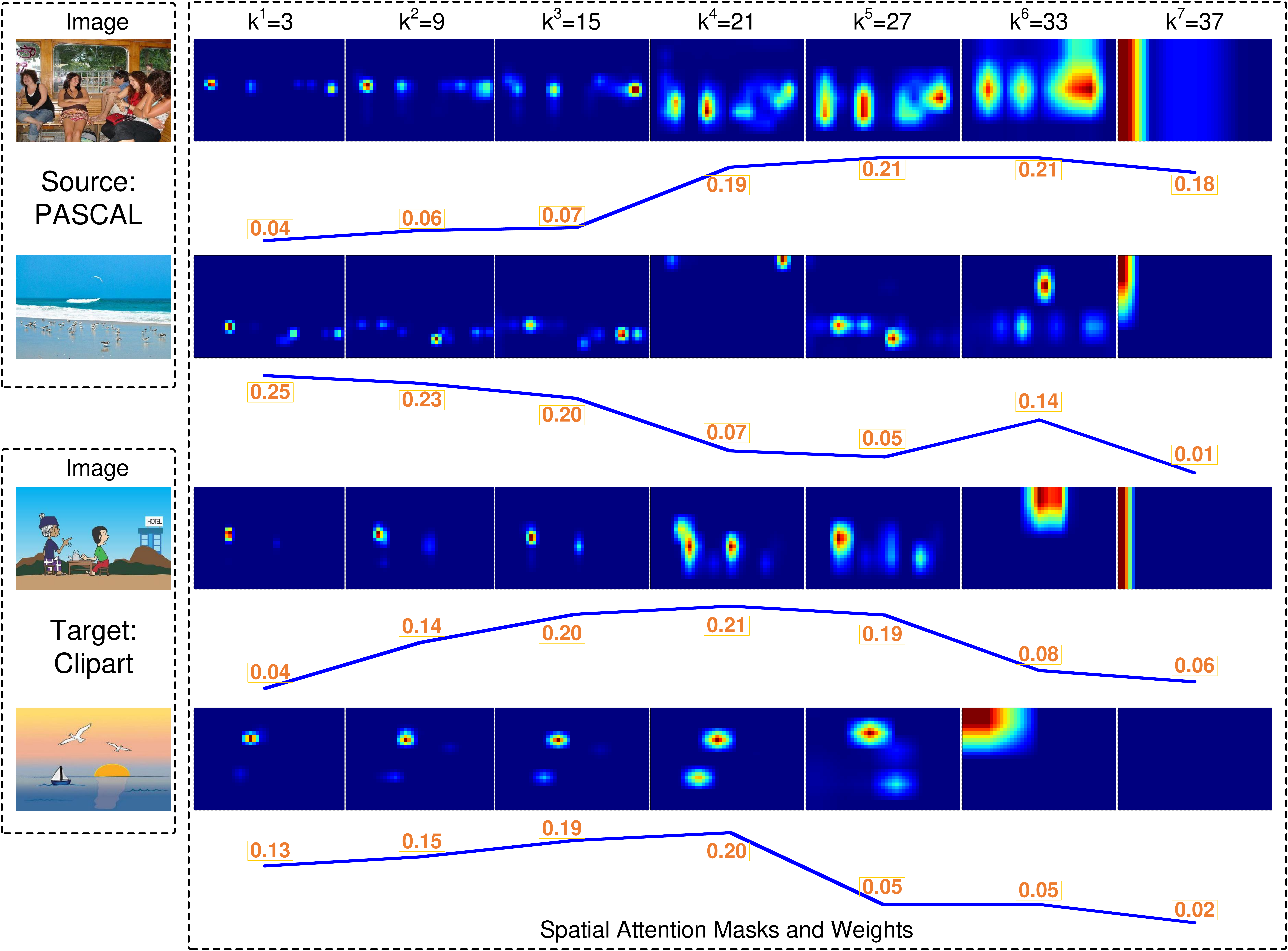}
\caption{Visualization of spatial attention masks and corresponding weights (blue lines) for the spatial pyramid. The attention masks of different feature maps are resized by the same scale for better visualization.}
\label{fig:spatial}
\end{figure}

Then, we concatenate the guided map $\mathcal{\hat{P}}$ and feature map $\hat{f}$ to generate guided feature map $\mathcal{\bar{P}}$ followed by $3$ convolutional layers. The guided feature map $\mathcal{\bar{P}}$ is shared for all $N$ scales. To adjust each scale of feature map $f^n$ in the spatial pyramid, we resize $\mathcal{\bar{P}}$ to the size $C\times H_{n} \times W_n$ at each level. The spatial attention mask $\mathcal{P}^n \in \mathbb{R}^{H_n \times W_n}$ can be predicted by the followed $3$ convolutional layers. Finally, $\mathcal{P}^n$ is normalized using the softmax function to compute the spatial attention, \ie,
\begin{equation}
\omega^{n}(x, y) = \frac{e^{\mathcal{P}^n(x, y)}}{\sum_{i=1}^{H_n}\sum_{j=1}^{W_n} e^{\mathcal{P}^n(i, j)}},
\end{equation}
where $\omega^{n}(i, j)$ indicates the value of the attention mask $\omega^{n}$ at $(i, j)$. Thus, we have $\sum_{i=1}^{H_n}\sum_{j=1}^{W_n}\omega^n(i, j)=1$.
As shown in Fig. \ref{fig:spatial}, we provide some examples with normalized attention masks $\omega^n$ for different feature maps in the spatial pyramid when $N = 7$. We can conclude that the feature map with different scale $k^n$ focuses on different semantic regions. For example, in the forth row, the feature map with smaller pooling size ($k=3$) pays more attention on the seagull, while the feature map with larger pooling size ($k=21$) focuses on the sail boat and the neighbouring context. Based on different guided information, $\omega^n$ recalibrates spatial responses in feature map $f^n$ adaptively.

On the other hand, it can be seen that not all the attention masks correspond to meaningful regions (see the attention mask with pooling size $k=37$ in the forth row). Inspired by \cite{DBLP:conf/cvpr/LiW0019}, we develop a dynamic weight selection mechanism to adjust the channel-wise weight of feature maps in the pyramid adaptively. To consider feature maps with different size, we normalize $f^n$ to an attention vector $V^{n} \in \mathbb{R}^{C \times 1}$ using the corresponding spatial attention weight $\omega^n$ as:
\begin{equation}
V^n = \sum_{i=1}^{H_n}\sum_{j=1}^{W_n}{f^n \cdot \omega^n},
\end{equation}
where $i$ and $j$ enumerate all spatial positions of weighted feature map $f^n \cdot \omega^n$. Thus the attention vectors have the same size for all the feature maps in the pyramid. Given attention vectors $\{V^1,\dots,V^N\}$, we first fuse these vectors via an element-wise addition, \ie, $v=\sum_{n=1}^NV^n$. Then, a compact feature ${\bf z} \in \mathbb{R}^{d \times 1}$ is created to enable the guidance for adaptive selections by the batch normalization layer, where $d$ is the dimension of the compact feature ${\bf z}$, and we set it to $\frac{C}{2}$ in all experiments. After that, for each attention vector $V^n$, we compute the channel-wise attention weight $\phi^n \in \mathbb{R}^{C \times 1}$ as
\begin{equation}
\phi^n = \frac{e^{a_n\cdot{\bf z}}}{\sum_{i=1}^N{e^{a_i\cdot{\bf z}}}},
\end{equation}
where $\{a_i, \dots, a_N\}$ are learnable parameters of fully connected layers for each scale. We have $\sum_{c=1}^N{\phi^n(c)}=1$, where $\phi^n(c)$ is the $c$-th element of $\phi^n$. In Fig.~\ref{fig:spatial}, we show the corresponding weights of each feature map in the spatial pyramid. Specifically, we compute the mean of channel-wise attention weight $\phi^n$ for each scale in each image. Finally, the fused semantic vector ${\cal V}\in \mathbb{R}^{C \times 1}$ is obtained through the channel-wise attention weight as ${\cal V}=\sum_{n=1}^N{V^n \cdot\phi^n}$, where ${\cal V}$ is a highly embedded vector in the latent space that encodes the semantic information of different spatial locations, different channels and the relations among them.
\begin{figure}[t]
\centering
\includegraphics[width=0.75\linewidth]{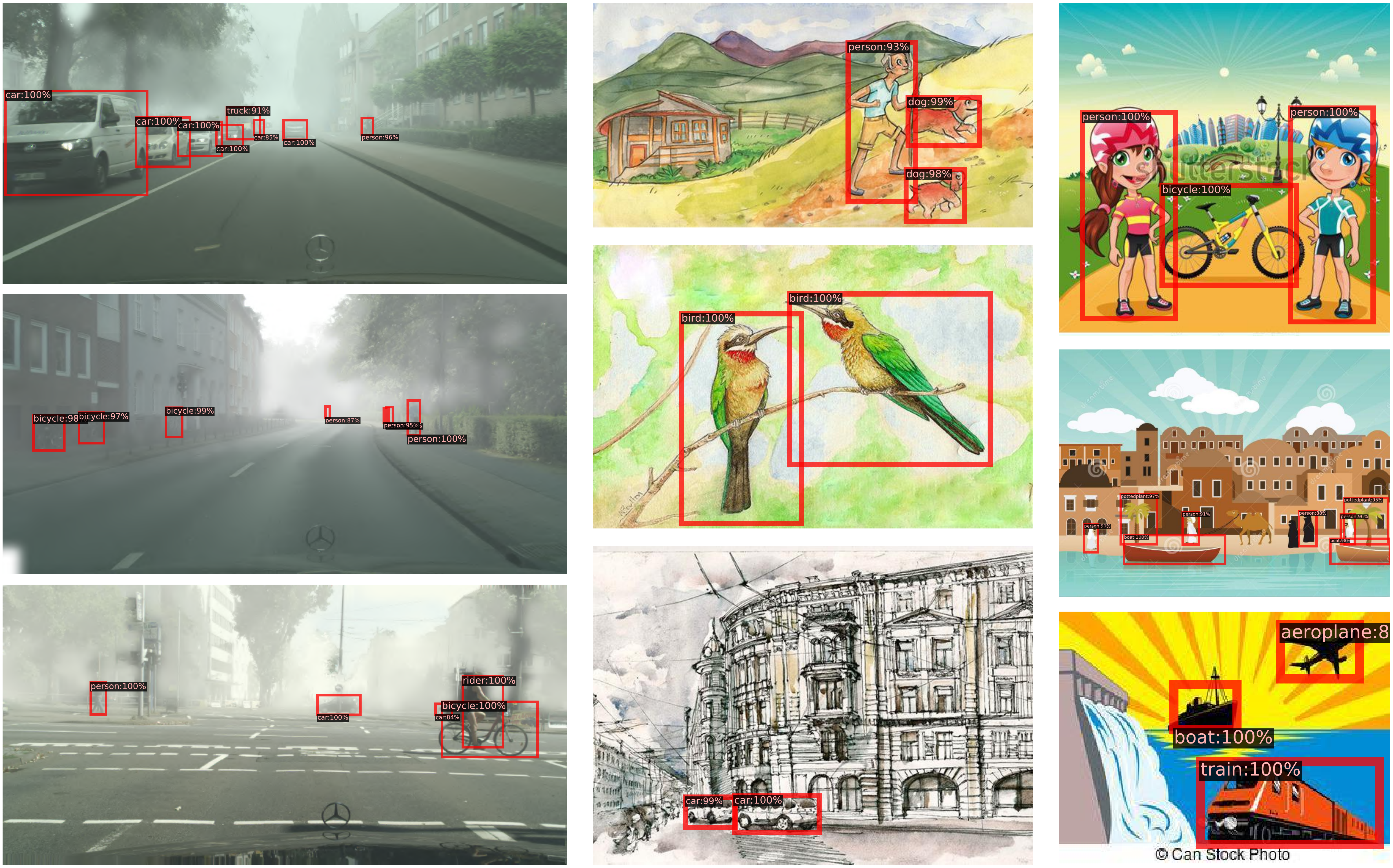}
\caption{Adaptation object detection results. From left to right: Foggy Cityscapes, Watercolor and Clipart.}
\label{fig:detection}
\end{figure}

{\flushleft {\bf Optimization}.}
The whole network is trained by minimizing two loss terms, \ie, adversarial loss and task-specific loss. The adversarial loss is used to determine whether the sample comes from the source domain or target domain. Specifically, we calculate the probability $x_i$ of the sample belonging to the target domain based on the fused semantic vector ${\cal V}$ using a simple fully-connected layer. The proposed SAPNet is denoted as $D$. Then, the adversarial loss is computed as
\begin{align}
\mathcal{L}_{\text{adv}}(G, D) =\frac{1}{|\mathcal{D}_s \cup \mathcal{D}_t|} \sum_{x_i \in \mathcal{D}_s \cup \mathcal{D}_t}\mathcal{H} (D(G(x_i)), y_i),
\end{align}
where $y_i$ is the domain label ($0$ for source domain and $1$ for target domain) and $\mathcal{H}$ is the cross-entropy loss function. On the other hand, the task loss $\mathcal{L}_{\text{task}}$ is determined by the specific task, \ie, object detection, instance segmentation, and semantic segmentation. The loss is computed as
\begin{align}
\mathcal{L}_{\text{task}}(G, R) =\frac{1}{|\mathcal{D}_s|} \sum_{x_i \in \mathcal{D}_s}\mathcal{L}_{\text{task-specific}}(R(G(x_i)), y_i^s),
\end{align}
where $G$ and $R$ are the backbone and task-specific components of the network, respectively. $y_i^s$ is the ground-truth label of sample $i$ in the source domain. We have $\mathcal{L}_{\text{task-specific}}=\{\mathcal{L}_{\text{det}},\mathcal{L}_{\text{ins}},\mathcal{L}_{\text{seg}}\}$. Taking object detection as an example, we denote the objective of Faster R-CNN as $\mathcal{L}_{\text{det}}$, which contains classification loss of object categories and regression loss of object bounding boxes. In summary, the overall objective is formulated as
\begin{align}
\max_{D}\min_{G, R}\mathcal{L}_{\text{task}}(G, R) - \lambda\mathcal{L}_{\text{adv}}(G, D),
\label{euq:loss}
\end{align}
where $\lambda$ controls the trade-off between task-specific loss and adversarial training loss. Following \cite{DBLP:conf/cvpr/Chen0SDG18,DBLP:conf/cvpr/SaitoUHS19}, we use gradient reverse layer (GRL) \cite{DBLP:conf/icml/GaninL15} to enable adversarial training where the gradient is reversed before back-propagating to $G$ from $D$. We first train the networks with only source domain to avoid initially noisy predictions. Then we train the whole model with Adam optimizer and the initial learning rate is set to $10^{-5}$, then divided by $10$ at $70,000$, $80,000$ iterations. The total number of training iterations is $90,000$.

\section{Experiment}
We implement our SAPNet method with PyTorch \cite{paszke2017automatic}, which is evaluated in three domain adaptation tasks, including object detection, instance segmentation, and semantic segmentation. For fair comparison, we set the shorter side of the image to $600$ following the implementation of \cite{DBLP:conf/cvpr/SaitoUHS19,DBLP:journals/corr/abs-1911-02559} with RoIAlign \cite{DBLP:conf/iccv/HeGDG17} in object detection; for instance segmentation and semantic segmentation, we use the same settings as previous methods. To consider the trade-off between accuracy and complexity, the number of pyramid levels is set to $N = 13$ for object detection and instance segmentation, \ie, we have the spatial pooling size set $K=\{3, 6, 9, 12, 15, 18, 21, 24, 27, 30, 33, 35, 37\}$. Note that we start from the initial pooling size $3\times3$ with the step of $3$, and the last two pooling sizes are reduced from $\{38, 41\}$ to $\{35, 37\}$ because of the width limit of feature map. For semantic segmentation, the number of pyramid levels is set to $N = 9$ since semantic segmentation involves feature maps with higher resolution, \ie, $K=\{3, 9, 15, 21, 27, 33, 39, 45, 51\}$. The hyper-parameter $\lambda$ is used to control the adaptation between source and target domains. Thus, we use different $\lambda$ in different tasks. Empirically, we set a larger $\lambda$ for adaptation between similar domains (\eg, Cityscapes$\to$FoggyCityscapes), and set a smaller $\lambda$ for adaptation between dissimilar domains (\eg, PASCAL VOC$\to$WaterColor). We choose $\lambda$ based on the performance on the validation set.
\begin{table*}[t]
\centering
\caption{Adaptation detection results from Cityscapes to FoggyCityscapes.}
\setlength{\tabcolsep}{2.5pt}
\begin{tabular}{l|cccccccc|c}
\hline
Method  &person &rider &car &truck &bus &train &cycle &bicycle &mAP \\
\hline
Faster R-CNN (w/o)     & 24.1& 33.1& 34.3& 4.1 & 22.3& 3.0 & 15.3& 26.5&  20.3 \\
DA-Faster \cite{DBLP:conf/cvpr/Chen0SDG18}& 25.0& 31.0& 40.5& 22.1& 35.3& 20.2& 20.0& 27.1&  27.6 \\
SCDA \cite{DBLP:conf/cvpr/ZhuPYSL19}&33.5 & 38.0 & 48.5 &26.5 &39.0 &23.3 &28.0 &33.6 &33.8 \\
Strong-Weak \cite{DBLP:conf/cvpr/SaitoUHS19} & 29.9& 42.3& 43.5& 24.5& 36.2& 32.6& 30.0& 35.3&  34.3 \\
Diversify\&match \cite{DBLP:conf/cvpr/KimJKCK19}& 30.8& 40.5& 44.3& 27.2& 38.4& 34.5& 28.4& 32.2&  34.6 \\
MAF \cite{DBLP:journals/corr/abs-1907-10343}         & 28.2& 39.5& 43.9& 23.8& 39.9& 33.3& 29.2& 33.9& 34.0\\
SCL \cite{DBLP:journals/corr/abs-1911-02559}         & 31.6& 44.0& 44.8& {\bf 30.4}& 41.8& {\bf 40.7}& {\bf 33.6}& 36.2&  37.9 \\
\hline
SAPNet                          &{\bf 40.8} &{\bf 46.7} &{\bf 59.8} &24.3 &{\bf 46.8} &37.5 &30.4 &{\bf 40.7} &{\bf 40.9} \\
\hline
\end{tabular}
\label{tab:city2foggy}
\end{table*}

\subsection{Domain Adaptation for Detection}
For object detection task, we conduct our experiments in $3$ domain shift scenarios: (1) similar domains; (2) dissimilar domains; and (3) from synthetic to real images. We compare our model to the state-of-the-art methods on $6$ domain shift datasets: Cityscapes \cite{DBLP:conf/cvpr/CordtsORREBFRS16} to FoggyCityscapes \cite{DBLP:journals/ijcv/SakaridisDG18}, Cityscapes \cite{DBLP:conf/cvpr/CordtsORREBFRS16} to KITTI \cite{DBLP:conf/cvpr/GeigerLU12}, KITTI \cite{DBLP:conf/cvpr/GeigerLU12} to Cityscapes \cite{DBLP:conf/cvpr/CordtsORREBFRS16}, PASCAL VOC \cite{DBLP:journals/ijcv/EveringhamGWWZ10} to Clipart \cite{DBLP:conf/cvpr/InoueFYA18}, PASCAL VOC \cite{DBLP:journals/ijcv/EveringhamGWWZ10} to Watercolor \cite{DBLP:conf/cvpr/InoueFYA18}, Sim10K \cite{DBLP:conf/icra/Johnson-Roberson17} to Cityscapes \cite{DBLP:conf/cvpr/CordtsORREBFRS16}. For fair comparison, we use ResNet101 and VGG-16 as the backbone and the last convolutional layer to enable domain adaptation as similar as that in \cite{DBLP:conf/cvpr/SaitoUHS19,DBLP:journals/corr/abs-1911-02559}. Some qualitative adaptation results of object detection are shown in Fig. \ref{fig:detection}.

{\flushleft \textbf{Cityscapes$\to$FoggyCityscapes.}} Notably, we evaluate our model between Cityscapes \cite{DBLP:conf/cvpr/CordtsORREBFRS16} and FoggyCityscapes \cite{DBLP:journals/ijcv/SakaridisDG18} (simulated attenuation coefficient $\beta=0.02$) at the most difficult level. Specifically, Cityscapes is the source domain, while the target domain FoggyCityscape (Foggy for short) is rendered from the same images in Cityscape using depth information. We set $\lambda = 1.0$ in \eqref{euq:loss} empirically. As shown in Table \ref{tab:city2foggy}, our SAPNet gains $3.0\%$ average accuracy improvement compared with the previous state-of-the-art methods. Specifically, in terms of \textit{person} and \textit{car} categories, our method outperforms the second performer with a huge margin (about $9\%$ and $15\%$ higher, respectively).

{\flushleft \textbf{Cityscapes$\leftrightarrow$KITTI.}} As shown in Table \ref{table:kitti2city}, we present the comparison between our model and state-of-the-art on domain adaptation between Cityscapes \cite{DBLP:conf/cvpr/CordtsORREBFRS16} and KITTI \cite{DBLP:conf/cvpr/GeigerLU12}. Similar to the works in \cite{DBLP:conf/cvpr/Chen0SDG18,DBLP:journals/corr/abs-1911-02559}, we use KITTI training set that contains $7,481$ images. We set $\lambda = 0.01$ for Cityscapes$\,\to\,$KITTI and $\lambda = 0.2$ for KITTI$\,\to\,$Cityscapes in \eqref{euq:loss} empirically. The Strong-Weak and SCL \cite{DBLP:journals/corr/abs-1911-02559} methods only employ multi-stage feature maps from the backbone to align holistic features, resulting in inferior performance than our method on both directions. In summary, our method achieves $1.5\%$ and $2.5\%$ accuracy improvement of KITTI$\,\to\,$Cityscapes and Cityscapes$\,\to\,$KITTI, respectively.

\begin{table*}[t]
\centering
\caption{Adaptation detection results between KITTI and Cityscapes. We report AP scores in terms of the car category on both directions, including KITTI$\,\to\,$Cityscapes and Cityscapes$\,\to\,$KITTI.}
\setlength{\tabcolsep}{9.5pt}
\begin{tabular}{l|c|c}
\hline
Method &KITTI$\,\to\,$Cityscapes &Cityscapes$\,\to\,$KITTI \\
\hline
Faster RCNN                              & 30.2    &  53.5    \\
DA-Faster \cite{DBLP:conf/cvpr/Chen0SDG18}           & 38.5    &  64.1    \\
Strong-Weak (impl. of \cite{DBLP:journals/corr/abs-1911-02559}) & 37.9   &  71.0   \\
SCL \cite{DBLP:journals/corr/abs-1911-02559}                   & 41.9    &  72.7    \\
SCDA \cite{DBLP:conf/cvpr/ZhuPYSL19}  & 42.5 & -\\
SAPNet                                     &{\bf 43.4} &{\bf 75.2} \\
\hline
\end{tabular}
\label{table:kitti2city}
\end{table*}

{\flushleft \textbf{PASCAL VOC$\to$Clipart/WaterColor.}} Moreover, we evaluate our method on dissimilar domains, \ie, from real images to artistic images. According to \cite{DBLP:conf/cvpr/SaitoUHS19}, PASCAL VOC \cite{DBLP:journals/ijcv/EveringhamGWWZ10} is the source domain, where the PASCAL VOC 2007 and 2012 training and validation sets are used for training. For the target domain, we use Clipart \cite{DBLP:conf/cvpr/InoueFYA18} and Watercolor \cite{DBLP:conf/cvpr/InoueFYA18} as that in \cite{DBLP:conf/cvpr/SaitoUHS19}. ResNet-101 \cite{DBLP:conf/cvpr/HeZRS16} pre-trained on ImageNet \cite{DBLP:conf/cvpr/DengDSLL009} is used as the backbone network following \cite{DBLP:conf/cvpr/SaitoUHS19,DBLP:journals/corr/abs-1911-02559}. We set $\lambda = 0.1$ and $\lambda = 0.01$ for Clipart \cite{DBLP:conf/cvpr/InoueFYA18} and Watercolor \cite{DBLP:conf/cvpr/InoueFYA18} respectively. As shown in Table \ref{table:voc2clipart} and Table \ref{table:voc2watercolor}, our model obtains comparable results with SCL \cite{DBLP:journals/corr/abs-1911-02559}.
\begin{table*}[t]
\centering
\caption{Adaptation detection results from PASCAL VOC to Clipart.}
\begin{tabular}{c|cccccccccc|c}
\hline
Method&aero&bicycle&bird&boat&bottle&bus&car&cat&chair&cow&\\
\hline
FRCNN \cite{DBLP:journals/pami/RenHG017}&35.6& 52.5& 24.3& 23.0& 20.0& 43.9& 32.8& 10.7& 30.6& 11.7& \\
BDC-Faster \cite{DBLP:conf/cvpr/SaitoUHS19}&20.2& 46.4& 20.4& 19.3& 18.7& 41.3& 26.5& 6.4& 33.2& 11.7&\\
DA-Faster \cite{DBLP:conf/cvpr/Chen0SDG18}&15.0& 34.6& 12.4& 11.9& 19.8& 21.1& 23.2& 3.1& 22.1& 26.3&\\
WST-BSR \cite{DBLP:journals/corr/abs-1909-00597}&28.0& 64.5& 23.9& 19.0& 21.9& \bf64.3& 43.5& 16.4& 42.2& 25.9& \\
Strong-Weak \cite{DBLP:conf/cvpr/SaitoUHS19}& 26.2& 48.5& 32.6& \bf33.7& 38.5& 54.3& 37.1& 18.6& 34.8& 58.3&\\
SCL \cite{DBLP:journals/corr/abs-1911-02559}&\bf44.7& 50.0& \bf33.6& 27.4& 42.2& 55.6& 38.3& \bf19.2& 37.9& \bf69.0& \\
Ours   &27.4&\bf70.8& 32.0& 27.9& \bf42.4& 63.5& \bf47.5&14.3&\bf48.2&46.1&\\
\hline
\hline
&table&dog&horse&bike&person&plant&sheep&sofa&train&tv&mAP\\
FRCNN \cite{DBLP:journals/pami/RenHG017}&13.8& 6.0& 36.8& 45.9& 48.7& 41.9& 16.5& 7.3& 22.9& 32.0& 27.8 \\
BDC-Faster \cite{DBLP:conf/cvpr/SaitoUHS19}&26.0& 1.7& 36.6& 41.5& 37.7& 44.5& 10.6& 20.4& 33.3& 15.5& 25.6  \\
DA-Faster \cite{DBLP:conf/cvpr/Chen0SDG18}&10.6& 10.0& 19.6& 39.4& 34.6& 29.3& 1.0& 17.1& 19.7& 24.8&19.8  \\
WST-BSR \cite{DBLP:journals/corr/abs-1909-00597}&30.5& 7.9& 25.5& 67.6& 54.5& 36.4& 10.3& 31.2& \bf57.4& 43.5& 35.7\\
Strong-Weak \cite{DBLP:conf/cvpr/SaitoUHS19}&17.0& 12.5& 33.8& 65.5& 61.6& 52.0& 9.3& 24.9& 54.1& 49.1&38.1 \\
SCL \cite{DBLP:journals/corr/abs-1911-02559}&30.1& \bf26.3& 34.4& 67.3& 61.0& 47.9& \bf21.4& \bf26.3& 50.1& 47.3&41.5 \\
Ours   &\bf31.8&17.9&\bf43.8 &\bf68.0&\bf68.1&\bf49.0&18.7&20.4&55.8&\bf51.3&\bf42.2 \\
\hline
\end{tabular}
\label{table:voc2clipart}
\end{table*}

\begin{table*}[t]
\centering
\caption{Adaptation detection results from PASCAL VOC to WaterColor.}
\setlength{\tabcolsep}{7.5pt}
\begin{tabular}{l|cccccc|c}
\hline
Method & bike &bird &car &cat &dog &person& mAP\\
\hline
Faster RCNN              & 68.8& 46.8& 37.2& 32.7& 21.3& 60.7& 44.6 \\
DA-Faster \cite{DBLP:conf/cvpr/Chen0SDG18} & 75.2& 40.6& 48.0& 31.5& 20.6& 60.0& 46.0 \\
Strong-Weak \cite{DBLP:conf/cvpr/SaitoUHS19} & \bf82.3& \bf55.9& 46.5& 32.7& 35.5& 66.7& 53.3 \\
SCL \cite{DBLP:journals/corr/abs-1911-02559}                 & 82.2& 55.1& 51.8& \bf39.6&38.4 & 64.0& \bf55.2 \\
Ours                          &81.1   &51.1 &\bf53.6 &34.3 &\bf39.8 &\bf71.3 & \bf55.2 \\
\hline
\end{tabular}
\label{table:voc2watercolor}
\end{table*}

\noindent \textbf{Sim10K$\to$Cityscapes.} In addition, we evaluate our model in the synthetic to real scenario. Following \cite{DBLP:conf/cvpr/Chen0SDG18,DBLP:conf/cvpr/SaitoUHS19}, we use Sim10K \cite{DBLP:conf/icra/Johnson-Roberson17} as the source domain that contains $10,000$ training images collected from the computer game Grand Theft Auto 5 (GTA5). We set $\lambda = 0.1$ in \eqref{euq:loss} empirically. As shown in Table \ref{table:sim102city}, our SAPNet obtains $3.3\%$ improvement in terms of AP score compared with state-of-the-art methods.

It is worth mentioning that BDC-Faster \cite{DBLP:conf/cvpr/SaitoUHS19} is also trained using cross-entropy loss but the performance is significantly decreased. Therefore, The Strong-Weak method \cite{DBLP:conf/cvpr/SaitoUHS19} adapts the focal loss \cite{DBLP:conf/iccv/LinGGHD17} to balance different regions. Compared with \cite{DBLP:conf/cvpr/SaitoUHS19,DBLP:journals/corr/abs-1911-02559}, our proposed attention mechanism is much more effective and thus the focal loss module is no longer needed.

\begin{table*}[t]
\begin{minipage}{0.49\linewidth}
\centering
\caption{Adaptation detection results from Sim10k to Cityscapes.}
\setlength{\tabcolsep}{9.5pt}
\begin{tabular}{l|c}
\hline\noalign{\smallskip}
Method & AP on Car \\
\hline
Faster R-CNN                   & 34.6    \\
DA-Faster \cite{DBLP:conf/cvpr/Chen0SDG18}& 38.9    \\
Strong-Weak \cite{DBLP:conf/cvpr/SaitoUHS19} & 40.1    \\
SCL \cite{DBLP:journals/corr/abs-1911-02559}         & 42.6    \\
SCDA \cite{DBLP:conf/cvpr/ZhuPYSL19}  & 43.0 \\
Ours                          & {\bf 44.9}    \\
\hline
\end{tabular}
\label{table:sim102city}
\end{minipage}
\begin{minipage}{0.49\linewidth}
\centering
\caption{Adaptation instance segmentation results from Cityscapes to FoggyCityscapes.}
\setlength{\tabcolsep}{9.5pt}
\begin{tabular}{l|c}
\hline\noalign{\smallskip}
Method & mAP \\
\hline
Source Only                   & 26.6    \\
SCDA \cite{DBLP:conf/cvpr/ZhuPYSL19}& 31.4    \\
Ours                          &{\bf 39.4}    \\
\hline
\end{tabular}
\label{table:city2foggymask}
\end{minipage}
\end{table*}

\subsection{Domain Adaptation for Segmentation}
\begin{figure}[t]
\centering
\includegraphics[width=.75\linewidth]{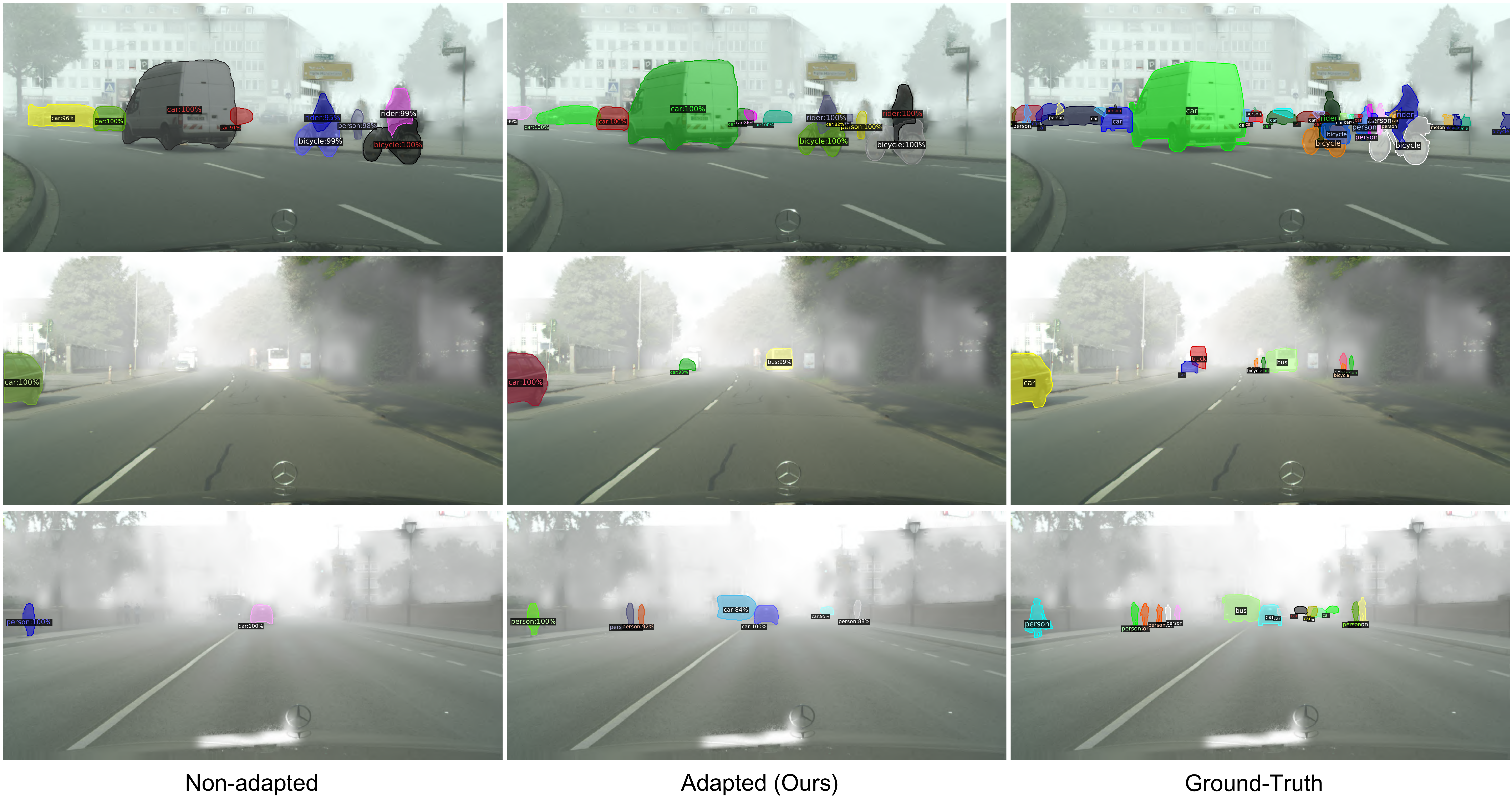}
\caption{Instance segmentation results for Cityscapes $\to$ Foggy Cityscapes.}
\label{fig:instance}
\end{figure}

\noindent \textbf{Instance Segmentation.}
For instance segmentation task, we evaluate our model from Cityscapes \cite{DBLP:conf/cvpr/CordtsORREBFRS16} to FoggyCityscapes \cite{DBLP:journals/ijcv/SakaridisDG18}. Similar to \cite{DBLP:conf/cvpr/ZhuPYSL19}, we use the VGG16 as the backbone network and add the segmentation head similar to that in Mask R-CNN \cite{DBLP:conf/iccv/HeGDG17}. From Table \ref{table:city2foggymask}, we can conclude that our method outperforms SCDA \cite{DBLP:conf/cvpr/ZhuPYSL19} significantly, \ie, $39.4$ vs. $31.4$. Some visual examples of adaptation instance segmentation results are shown in Fig. \ref{fig:instance}.

\begin{figure}[t]
\centering
\includegraphics[width=.75\linewidth]{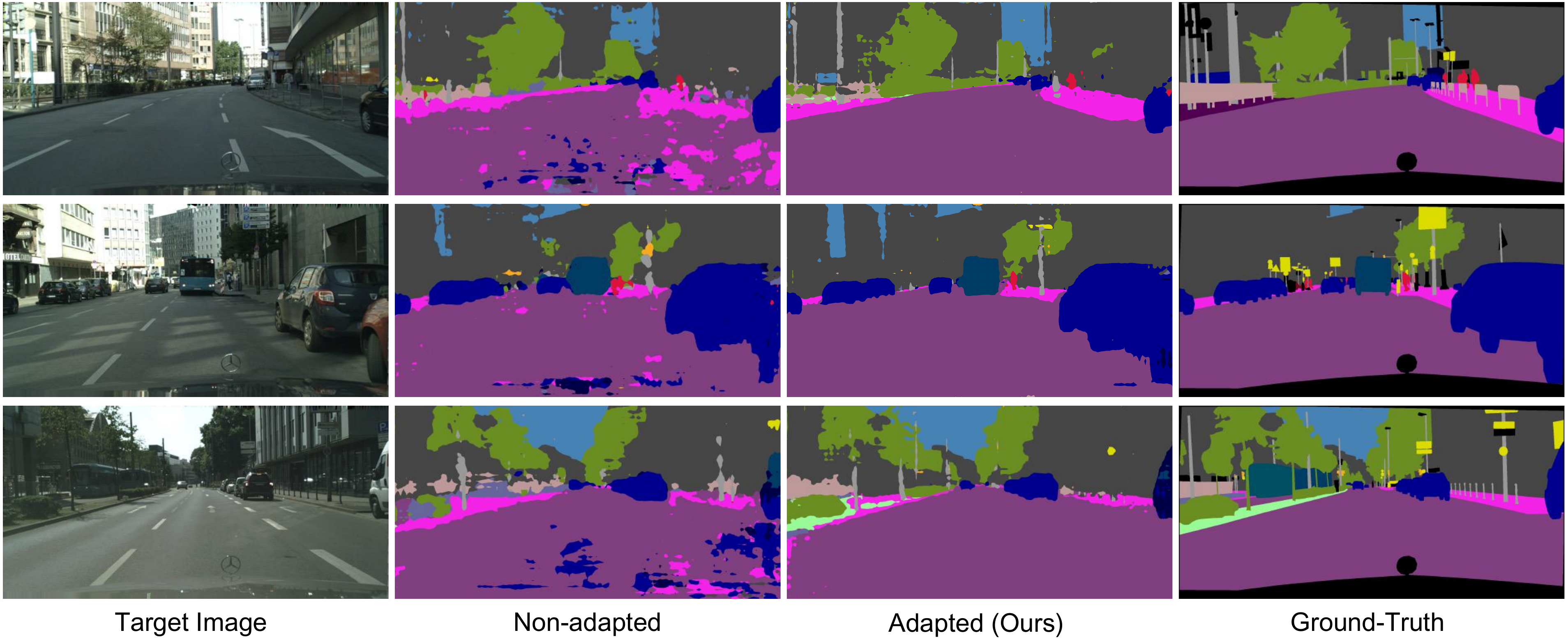}
\caption{Semantic segmentation results for GTA5 $\to$ Cityscapes.}
\label{fig:semantic}
\end{figure}

\noindent \textbf{Semantic Segmentation.}
For semantic segmentation task, we conduct experiments from GTA5 \cite{DBLP:conf/eccv/RichterVRK16} to Cityscapes \cite{DBLP:conf/cvpr/CordtsORREBFRS16} and SYNTHIA \cite{DBLP:conf/cvpr/RosSMVL16} to Cityscapes. Following \cite{DBLP:conf/cvpr/Luo0GYY19}, we use the DeepLab-v2 \cite{DBLP:journals/pami/ChenPKMY18} framework with ResNet-101 backbone that is pre-trained on ImageNet. Notably, the task-specific guided map for semantic segmentation naturally comes from the predicted output with the shape of $C_{sem}\times H\times W$, where $C_{sem}$ is the number of semantic categories. As presented in Table \ref{table:gta2cityscapes} and Table \ref{table:synthia2cityscapes}, our method achieves comparable segmentation accuracy with state-of-the-arts on the domain adaptation from GTA5 \cite{DBLP:conf/eccv/RichterVRK16} to Cityscapes \cite{DBLP:conf/cvpr/CordtsORREBFRS16}, and from SYNTHIA \cite{DBLP:conf/cvpr/RosSMVL16} to Cityscapes. Some visual examples of adaptation semantic segmentation results are shown in Fig. \ref{fig:semantic}.
\begin{table*}[t]
\centering
\caption{Adaptation semantic segmentation results from GTA5 to Cityscapes.}
\setlength{\tabcolsep}{3.0pt}
\begin{tabular}{l|ccccccccccccccccccc|c}
\hline
Method & road &side &buil. &wall &fence &pole &light &sign &vege. &terr. \\
\hline
Source &75.8 &16.8 &77.2 &12.5 &21.0 &25.5 &30.1 &20.1 &81.3 &24.6  \\
ROAD \cite{DBLP:conf/cvpr/Chen0G18} &76.3 &36.1 &69.6 &28.6 &22.4 &\bf28.6 &29.3 &14.8 &82.3 &\bf35.3 \\
TAN \cite{DBLP:conf/cvpr/TsaiHSS0C18} &86.5 &25.9 &\bf79.8 &22.1 &20.0 &23.6 &33.1 &21.8 &81.8 &25.9  \\
CLAN \cite{DBLP:conf/cvpr/Luo0GYY19} &87.0 &27.1 &79.6 &27.3 &23.3 &28.3 &\bf35.5 &\bf24.2 &\bf83.6 &27.4  \\
Ours &{\bf 88.4} &{\bf 38.7} &79.5 &{\bf 29.4} &{\bf 24.7} &27.3 & 32.6& 20.4 &82.2 &32.9 \\
\hline
\hline
&sky  &pers. &rider &car &truck &bus &train &motor &bike &mIoU \\
Source &70.3 &53.8 &26.4 &49.9 &17.2 &25.9 &6.5 &25.3 &\bf36.0 &36.6 \\
ROAD \cite{DBLP:conf/cvpr/Chen0G18} &72.9 &54.4 &17.8 &78.9 &27.7 &30.3 &4.0 &24.9 &12.6 &39.4 \\
TAN \cite{DBLP:conf/cvpr/TsaiHSS0C18} &\bf75.9 &57.3 &26.2 &76.3 &29.8 &32.1 &\bf7.2 &29.5 &32.5 &41.4 \\
CLAN \cite{DBLP:conf/cvpr/Luo0GYY19} &74.2 &\bf58.6 &\bf28.0 &76.2 &\bf33.1 &36.7 &6.7 &\bf31.9 &31.4 &43.2 \\
Ours &73.3 &55.5 &26.9 &{\bf 82.4} &31.8 &{\bf 41.8} &2.4 &26.5 &24.1 &{\bf 43.2} \\
\hline
\end{tabular}
\label{table:gta2cityscapes}
\end{table*}

\begin{table*}[t]
\centering
\caption{Adaptation semantic segmentation results from SYNTHIA to Cityscapes.}
\setlength{\tabcolsep}{1.5pt}
\begin{tabular}{l|ccccccccccccc|c}
\hline
Method & road &side &buil. &light &sign &vege. &sky &pers. &rider &car &bus &motor &bike &mIoU\\
\hline
Source &55.6 &23.8 &74.6 &6.1 &12.1 &74.8 &79.0 &55.3 &19.1 &39.6 &23.3 &13.7 &25.0 &38.6 \\
TAN \cite{DBLP:conf/cvpr/TsaiHSS0C18} &79.2 &37.2 &78.8 &9.9 &10.5 &78.2 &80.5 &53.5 &19.6 &67.0 &29.5 &21.6 &31.3 &45.9 \\
CLAN \cite{DBLP:conf/cvpr/Luo0GYY19} &81.3 &37.0 &80.1 &16.1 &13.7 &78.2 &81.5 &53.4 &21.2 &73.0 &32.9 &22.6 &30.7 &47.8 \\
Ours &81.7 &33.5 &75.9 &7.0 &6.3 &74.8 &78.9 &52.1 &21.3 &75.7 &30.6 &10.8 &28.0 &44.3 \\
\hline
\end{tabular}
\label{table:synthia2cityscapes}
\end{table*}

\subsection{Ablation Study}
We further perform experiments to study the effect of important aspects in SAPNet, \ie, task-specific guided map and spatial attention pyramid. Since PASCAL VOC $\to$ Clipart, Sim10k $\to$ Cityscapes and Cityscapes $\to$ Foggy represent three different domain shift scenarios, we perform ablation study in terms of object detection datasets for comprehensive analysis.

{\noindent {\bf Task-specific guided information}:} To investigate the importance of task-specific guided information, we remove the task-specific guidance to generate the spatial attention mask, which is denoted as ``w/o GM''. In this way, the number of channels of the first convolutional layer after concatenation of feature maps in layer 3 and task-specific guided information is reduced (see Fig. \ref{fig:framework}). However, the impact is negligible since the channel number of guided map is small. As presented in Table \ref{table:ablation}, the task-specific guided information improves the accuracy, especially for dissimilar domains PASCAL VOC and Clipart ($42.2$ vs. $37.1$). We speculate that such guidance can facilitate focusing on the most discriminative semantic regions for domain adaptation.

{\noindent {\bf Spatial attention pyramid}:} To investigate the effectiveness of spatial attention pyramid, we construct the ``w/o SA'' variant of SAPNet, which indicates that we remove the spatial attention masks and global attention pyramid (since no multi-scale vectors are available) in Fig. \ref{fig:framework}. As shown in Table \ref{table:ablation}, the performance drops dramatically without spatial attention pyramid.  On the other hand, along with the increasing number of pooled feature maps in the pyramid, the performance is gradually improved. Specifically, we use the spatial pooling size set $K=\{3, 6, 9, 12, 15, 18, 21, 24, 27, 30, 33, 35, 37\}$ when $N=13$, $K=\{3, 9, 15, 21, 27, 33, 37\}$ when $N=7$ and $K=\{3, 21, 37\}$ when $N=3$. It indicates that the spatial pyramid with deep levels contains more discriminative semantic information for domain adaptation and our method can make full use of it. In addition, we compare average pooling and maximal pooling operations in spatial attention pyramid. We can conclude that average pooling achieves better performance in different datasets, which demonstrates the effectiveness of average pooling to capture discriminative local patterns for domain adaptation.
\begin{table*}[t]
\centering
\caption{Effectiveness of important aspects in SAPNet.}
\label{table:ablation}
\setlength{\tabcolsep}{3.5pt}
\begin{tabular}{c|c|c|c}
\hline
Variant & PASCAL VOC$\to$Clipart & Sim10k$\to$Cityscapes & Cityscapes$\to$Foggy \\
\hline
w/o GM &37.1 &43.8 &38.3  \\
w/ GM   &\bf42.2 &\bf44.9 &\bf40.9  \\
\hline\hline
w/o CA & 37.7 & 45.6 & 40.4 \\
w/ CA   & \bf42.2 & \bf44.9 & \bf40.9  \\
\hline\hline
w/o SA & 35.4 & 38.3 & 36.6  \\
w/ SA($N=3$)  & 39.6 & 43.9 & 39.0 \\
w/ SA($N=7$)  & 40.2 & \bf45.9 & 40.5 \\
w/ SA($N=13$) & \bf42.2 & 44.9 & \bf40.9 \\
\hline\hline
max pooling & 37.5 & 43.1 & 34.9 \\
avg pooling & \bf42.2 & \bf44.9 & \bf40.9 \\
\hline
\end{tabular}
\end{table*}

{\noindent {\bf Channel-wise attention}:}
To verify the effectiveness of channel-wise attention, we conduct two variants to compute the embedded vector $\mathcal{V}$, where weighted summation $\mathcal{V}=\sum_{n=1}^N{V^n \cdot w^n}$ and equal summation $\mathcal{V}=\frac{1}{N}\sum_{n=1}^N{V^n}$ are denoted as ``w/ CA'' and ``w/o CA'' respectively. The results are shown in Table~\ref{table:ablation}. Notably, for similar domains (\eg, Sim10k to Cityscapes or Cityscapes to Foggy Cityscapes), we obtain very similar result without channel-wise attention; while for dissimilar domains (\eg, PASCAL to Clipart or PASCAL to Watercolor), we observe an obvious drop in performance, \ie, $4.5\%$ vs. $2.9\%$. This is maybe because similar/dissimilar domains share similar/different semantic information in each feature map of the spatial pyramid.

\section{Conclusions}
In this work, we propose a general unsupervised domain adaptation framework for various computer vision tasks including object detection, instance segmentation and semantic segmentation. Given target-specific guided information, our method can make full use of feature maps in the spatial attention pyramid, which enforces the network to focus on the most discriminative semantic regions for domain adaptation. Extensive experiments conducted on various challenging domain adaptation datasets demonstrate the effectiveness of the proposed, which performs favorably against the state-of-the-art methods.

\section*{Acknowledgement}
This work was supported by the Key Research Program of Frontier Sciences, CAS, Grant No. ZDBS-LY-JSC038, the National Natural Science Foundation of China, Grant No. 61807033 and National Key Research and Development Program of China (2017YFB0801900). Libo Zhang was supported by Youth Innovation Promotion Association, CAS (2020111), and Outstanding Youth Scientist Project of ISCAS.

\end{document}